\documentclass[manuscript,screen]{acmart}

\AtBeginDocument{%
  \providecommand\BibTeX{{%
    \normalfont B\kern-0.5em{\scshape i\kern-0.25em b}\kern-0.8em\TeX}}}

\usepackage[T1]{fontenc}
\usepackage{lmodern}
\begin{document}

\title{HinFlair: a pre-trained contextual string embeddings for pos tagging and text classification in Hindi language}

\author{Harsh Patel}
\email{patel.harsh1014@gmail.com}
\orcid{0000-0002-1468-2650}
\affiliation{%
  \institution{Medi-Caps University}
  \streetaddress{AB Rd, Pigdamber, Rau}
  \city{Indore}
  \state{Madhya Pradesh}
  \country{India}
  \postcode{453331}
}


\begin{abstract}
Recent advancements in language models based on recurrent neural networks and transformers architecture have achieved state-of-the-art results on a wide range of natural language processing tasks such as pos tagging, named entity recognition, and text classification. However, most of these language models are pre-trained in high resource languages like English, German, Spanish. Multi-lingual language models include Indian languages like Hindi, Telugu, Bengali in their training corpus, but they often fail to represent the linguistic features of these languages as they are not the primary language of the study.
We introduce HinFlair, which is a language representation model (contextual string embeddings) pre-trained on a large monolingual Hindi corpus. Experiments were conducted on 6 text classification datasets and a Hindi dependency treebank to analyze the performance of these contextualized string embeddings for the Hindi language. Results show that HinFlair outperforms previous state-of-the-art publicly available pre-trained embeddings for downstream tasks like text classification and pos tagging. Also, HinFlair when combined with FastText embeddings outperforms many transformers based language model trained particularly for Hindi language.\\
The datasets and other resources used for this study are publicly available at \url{https://github.com/harshpatel1014/HinFlair}
\end{abstract}




\maketitle

\section{Introduction}
Different NLP tasks like pos tagging, named entity recognition, question answering, sentiment analysis, machine translation have seen significant improvements in recent years. Improved deep learning techniques and data availability has led to the development of many language representation model and embeddings. These language models and embeddings capture deep semantic/syntactic features of the language resulting in less dependence on feature engineering. These advancements have engendered the NLP community to transition from task-specific problems to fine-tuning models for downstream tasks. However, most of the models and embeddings are primarily trained in high resource languages, and few models are trained in low resource languages.
Languages spoken in India are diverse. Around 1.3 billion people in India communicate with these diverse languages, with Hindi being the most spoken (500 million). Therefore, there is a great need for language models and embeddings representing these languages for social, cultural, and linguistic reasons\cite{ruder2020beyondenglish}.\par
Word embeddings are important decisive components for general NLP tasks like sequence labeling, text classification, entity extraction, etc. Typically, there are three types of distinctive word embeddings. First, classic word embedding trained over large data that captures syntactic and semantic similarity like Word2Vec \cite{mikolov2013distributed}, GloVe \cite{pennington2014glove}. Second, embeddings that capture character level sub-word features like FastText \cite{bojanowski2017enriching}. Third, embeddings that address context dependency and polysemy of words like ELMo \cite{peters2018deep}.\par
The introduction of transformers based architecture replaced recurrent layers with multi-headed self attention \cite{vaswani2017attention}. Models based on RNN architecture are not effective in handling long-term dependencies and prevents parallelization even with attention mechanism \cite{hochreiter2001gradient, parikh2016decomposable}. Transformers seems to address these issues effectively and are therefore ideal when dealing with NLP tasks like machine translation and question answering. Researchers have since then introduced many transformers based architecture, with one capturing context beyond fixed length \cite{dai2019transformer}, other capturing context in both directions \cite{devlin2018bert}, while some incorporating the best of earlier models \cite{yang2019xlnet}, each giving state-of-the-art results for various NLP tasks.\par
However, there are few language representation models and pre-trained embeddings for low resource languages like Hindi. In this article, I introduce HinFlair, pre-trained embeddings for the Hindi language. HinFlair is based on Flair embeddings \cite{akbik2018coling}, which achieves state-of-art of results on various sequence labeling tasks. Flair embeddings are pre-trained contextual string embeddings that combine the best of different types of embeddings mentioned above. Same as flair embeddings, HinFlair is trained on a large monolingual Hindi corpus that captures character level contextualized features. I evaluated HinFlair embeddings on six text classification datasets and one pos tagging dataset, and results show that HinFlair significantly outperforms previous state-of-the-art embeddings and language models.

\section{Related Work}
There are limited literature and research work on language representation models and pre-trained word embeddings when it comes to the Hindi language. {\itshape AI4Bharat-IndicNLP} Corpus have trained FastText embedding on their Hindi monolingual corpus containing around 63 million sentences \cite{kunchukuttan2020ai4bharat}. Their pre-trained embeddings have given state-of-the-art results on text classification. {\itshape Indic-Transformers} have trained 4 different transformers based architecture for the Hindi language and have achieved the best results for various NLP tasks \cite{jain2020indic}. Likewise, there is a comprehensive study where English text classification datasets are converted to the Hindi language, and they tested performance of around 8 different neural architectures on these converted datasets \cite{joshi2019deep}. Models on {\itshape iNLTK} library also gives state-of-the-art result for text classification \cite{arora2020inltk}. In this study, I have compared performance of HinFlair with the results published from these researches stated above.

\section{Materials and methods}
The following sections describe the dataset used for training HinFlair embeddings, along with the details of pos tagging and text classification datasets used for evaluation of the embedding. Furthermore, technical details of training the embedding and fine-tuning are presented.
\subsection{Datasets}
HinFlair embedding is trained on a large monolingual Hindi corpus produced by IIT Bombay \cite{kunchukuttan-etal-2018-iit}. The monolingual corpus is created by collecting Hindi text from various sources containing total of around 45 million sentences. For this study, I have used 80 percent of the data for training, while 10 percent of data is used for validation and testing each. The statistics of the monolingual corpus are listed in Table~\ref{tab:monolingual}.

\begin{table}
  \caption{Statistics of Monolingual Hindi Data}
  \label{tab:monolingual}
  \begin{tabular}{ccl}
    \toprule
    Source&No. of Sentences\\
    \midrule
    BBC-new&18,098\\
    BBC-old&135,171\\
    HindMonoCorp&44,486,496\\
    Health Domain&8,001\\
    Tourism Domain&15,395\\
    Wikipedia&259,305\\
    Judicial Domain&152,776\\
    \midrule
    Total&45,075,242\\
    \bottomrule
\end{tabular}
\end{table}

\begin{table*}
  \caption{Statistics of Classification and POS tagging datasets}
  \label{tab:classification}
  \begin{tabular}{cccl}
    \toprule
    Dataset &No. of Classes &Train  &Test\\
    \midrule
    IITP Movie  &3  &2480   &310\\
    IITP Product    &3  &4182   &523\\
    BBC Articles    &14 &3468   &867\\
    Trec-6  &6   &5452   &500\\
    SST-1   &5   &8544   &2210\\
    SST-2   &2  &6920   &1821\\
    UD Hindi &30   &13304   &1684\\
    \bottomrule
  \end{tabular}
\end{table*}

HinFlair embeddings performance is tested on 6 text classification datasets: IIT Patna movie review dataset \cite{akhtar2016hybrid}, IIT Patna product review dataset \cite{akhtar2016hybrid}, BBC articles for text classification. All these datasets are in Hindi language. The other 3 datasets: Trec-6 question corpus \cite{voorhees2000overview}, Stanford Sentiment Datasets SST-1 and SST-2 are also used for evaluation of embeddings \cite{socher2013recursive}. These datasets are translated from their original versions in English to Hindi language using Google Translate. While Hindi Universal Dependency Treebank is used for testing HinFlair on pos tagging task \cite{bhat2017hindi, palmer2009hindi}. Table~\ref{tab:classification} shows statistics about these datasets.

\subsection{HinFlair Training Details}
\begin{figure}[h]
  \centering
  \includegraphics[scale=0.70]{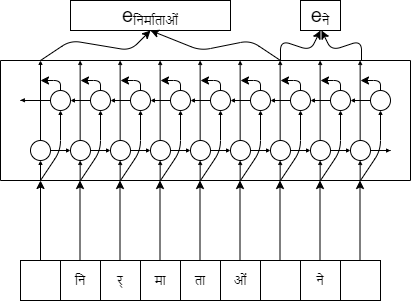}
  \caption{Overall approach for training HinFlair embeddings}
  \label{fig:1}
  \Description{Overall approach for training HinFlair embeddings}
\end{figure}

For this study, I have used state-of-the-art Flair embeddings architecture model to train HinFlair \cite{akbik2018coling}. HinFlair is trained on large monolingual corpus. The language model embedding is trained in the forward direction. Tokens from the corpus are fed as a sequence of characters into bidirectional LSTMs \cite{hochreiter1997long, graves2013generating}. Bidirectional LSTMs captures context from both direction flexibly encoding long-term dependencies better than typical RNNs \cite{jozefowicz2016exploring}. The output from both hidden states are concatenated after last character in the word, giving contextualized string embeddings for each word in a sequence \cite{akbik2019naacl}. The overall approach is illustrated in Figure~\ref{fig:1}.\par

Hindi language is written in Devanagari script. Therefore, a character dictionary from a monolingual corpus was created before training the model. Parameters for training the model are selected as per the recommendation from the authors of Flair. The hidden size of 1024 is taken for both LSTMs. Sequence length of 250 and a mini-batch size of 100 is selected. Model training is initialized with a learning rate of 20, annealed by the factor of 4 for every 25 splits with no improvement. HinFlair embeddings model was trained for 10 epochs for more than a week. The model reached the validation perplexity of 3.44 at end of the training.

\subsection{Experiments}
\textbf{POS tagging.} POS tagging is a sequence labeling problem. Popular neural network architecture BiLSTMs-CRF is employed on top of HinFlair embeddings in this experiment \cite{huang2015bidirectional}. Flair NLP library allows concatenating different word embeddings together. Experiments on high resource languages show that the model gives better results when Flair embedding is combined with classical word embeddings \cite{akbik2018coling}. Therefore, for this experiment, I have concatenated HinFlair embeddings with FastText embeddings trained in the Hindi language. Model is trained with an initial learning rate of 0.1 for 200 epochs. LSTMs hidden size is set to 256 with a batch size equal to 32. F1 score metric is used to test performance of the model.\\
\textbf{Text Classification.} The initial settings for text classification is same as pos tagging. Vector representations for each word from HinFlair embeddings and FastText embedding are concatenated together. These representations are taken as input to GRU network instead of BiLSTMs-CRF network \cite{chung2014empirical}. The output of GRU network is single embedding for complete sentence. The hidden size of GRU is 256. Other parameters are same as used for model training of pos tagging. Accuracy metric is used for evaluation of HinFlair for text classification.

\section{Results}
The results of HinFlair embeddings on pos tagging and 6 text classification datasets are listed in Table~\ref{tab:result1} and Table~\ref{tab:result2} respectively.

\begin{table}[t!]
  \caption{Test Results for POS tagging}
  \label{tab:result1}
  \begin{tabular}{cccc}
    \toprule
    Tags &Precision   &Recall  &F1-score\\
    \midrule
    PRP     &0.9903    &0.9896    &0.9900\\      
    PSP     &0.9972    &0.9983    &0.9978\\      
    NNPC     &0.9004    &0.9089    &0.9047\\      
    NNP     &0.9395    &0.9269    &0.9331\\      
    SYM     &1.0000    &1.0000    &1.0000\\      
    CC     &0.9915    &0.9930    &0.9922\\      
    RP     &0.9938    &0.9856    &0.9897\\   
    JJ     &0.9450    &0.9695    &0.9571\\      
    NN     &0.9733    &0.9702    &0.9718\\      
    VM     &0.9951    &0.9948    &0.9949\\      
    QF     &0.9576    &0.9679    &0.9627\\   
    VAUX     &0.9941    &0.9964    &0.9953\\      
    QC     &0.9899    &0.9933    &0.9916\\   
    NST     &0.9940    &0.9940    &0.9940\\      
    INTF     &0.8571    &0.9231    &0.8889\\      
    NNC     &0.8450    &0.8390    &0.8420\\      
    NEG     &0.9947    &0.9947    &0.9947\\      
    DEM     &0.9764    &0.9892    &0.9828\\      
    QO     &0.9815    &0.9298    &0.9550\\   
    RB     &0.9683    &0.9037    &0.9349\\
    RDP     &0.8750    &0.8750    &0.8750\\      
    JJC     &0.8235    &0.5833    &0.6829\\      
    WQ     &0.9545    &1.0000    &0.9767\\   
    QCC     &0.9694    &0.9596    &0.9645\\      
    PRPC     &1.0000    &0.7500    &0.8571\\
    UNK     &0.1875    &0.4286    &0.2609\\      
    NSTC     &1.0000    &1.0000    &1.000\\
    RBC     &1.0000    &0.0000    &0.0000\\      
    QFC     &1.0000    &0.0000    &0.0000\\      
    CCC     &1.0000    &1.0000    &1.0000\\
    \midrule
    F1-score    &   &   &97.44\\
    \bottomrule
  \end{tabular}
\end{table}

Results show that HinFlair achieves state-of-the-art results on 5 out of 6 text classification datasets. HinFlair achieves the best accuracy of 62.26 beyond 57.74 on IITP Movie, best accuracy of 77.25 beyond 75.71 on IITP product review dataset. On BBC article dataset, HinFlair gets the second-best score of 77.6. Results here are compared with transformers based models available on iNLTK. When compared with other word embeddings, HinFlair outperforms embeddings like INLP by at least 10 points on each dataset. For this study, I have converted three text classification datasets to Hindi language. HinFlair achieves a score of 94.39 on trec dataset, 40.7 on SST-1, and 78.74 on SST-2 dataset. There aren’t any pre-trained word embedding in the Hindi language that is tested on these datasets.\par
For POS tagging task, HinFlair gets the best F1-score of 97.44 on Universal Dependency Hindi treebank containing 30 different tags on pos tagging whose results are listed in Table~\ref{tab:result1}.

\begin{table}[t!]
  \caption{Test Results for Text Classification in Hindi}
  \label{tab:result2}
  \begin{tabular}{cccccc}
    \toprule
    Dataset &FT-W   &FT-WC  &INLP   &iNLTK  &HinFlair\\
    \midrule
    IITP Movie  &41.61   &44.52   &45.81   &57.74   &\textbf{62.26}\\
    IITP Product    &58.32   &57.17   &63.48   &75.71   &\textbf{77.25}\\
    BBC Articles    &72.29   &67.44   &74.25   &\textbf{78.75}   &77.6\\
    Trec-6  &-   &-   &-   &-   &\textbf{94.39}\\
    SST-1   &-   &-   &-   &-   &\textbf{40.7}\\
    SST-2   &-   &-   &-   &-   &\textbf{78.74}\\
    \bottomrule
  \end{tabular}
\end{table}

\section{Conclusion and Future Work}
This article presents HinFlair, pre-trained contextualized string embeddings for the Hindi language. Results show that HinFlair significantly outperforms previous word embeddings for NLP tasks like text classification and pos tagging. HinFlair embedding is trained in the forward direction. For future work, I plan to train HinFlair embedding on the same corpus but in a backward direction as combining embeddings trained in the opposite direction further improves the performance on various NLP tasks. Also, there aren’t many transformers models trained on a large Hindi corpus. Therefore, for future work, I would like to train language models based on transformers architecture for Indic languages.

\section*{acknowledgments}
I would like to thank the Department of Computer Science and Engineering from Medi-Caps University for assistance. I also thank anonymous reviewers for their suggestions and comments.
\bibliographystyle{ACM-Reference-Format}
\bibliography{ref}

\end{document}